\documentclass[11pt]{article}

\usepackage{arxiv}
\usepackage{authblk}

\usepackage{url}
\usepackage[square,sort,comma,numbers]{natbib}
\usepackage{hyperref}

\usepackage{times}
\usepackage{latexsym}

\usepackage[T1]{fontenc}

\usepackage[utf8]{inputenc}

\usepackage{microtype}

\usepackage{inconsolata}

\usepackage{graphicx}

\usepackage{amsmath}
\usepackage{amsfonts}

\usepackage{colortbl}
\usepackage[table]{xcolor}

\usepackage{float}
\usepackage{multirow}
\usepackage{tabularx}
\newcolumntype{L}[1]{>{\raggedright\arraybackslash}p{#1}}
\usepackage{booktabs}

\usepackage{enumitem}
\usepackage{xcolor}

\usepackage{algorithm}
\usepackage{algorithmicx}
\usepackage{algpseudocode}

\author[, 1]{Maxime Delmas\thanks{Corresponding author: maxime.delmas@idiap.ch}}
\author[1,2]{Lei Xu}
\author[1,3,4]{Andr\'{e} Freitas}
\affil[1]{Idiap Research Institute, Switzerland}
\affil[2]{École Polytechnique Fédérale de Lausanne (EPFL), Switzerland}
\affil[3]{Department of Computer Science, University of Manchester, United Kingdom}
\affil[4]{Cancer Biomarker Centre, CRUK Manchester Institute, United Kingdom}
\date{}
\title{A Navigational Approach for Comprehensive RAG via Traversal over Proposition Graphs}

\begin{document}
\maketitle
\begin{abstract}
Standard RAG pipelines based on chunking excel at simple factual retrieval but fail on complex multi-hop queries due to a lack of structural connectivity. Conversely, initial strategies that interleave retrieval with reasoning often lack global corpus awareness, while Knowledge Graph (KG)-based RAG performs strongly on complex multi-hop tasks but suffers on fact-oriented single-hop queries. To bridge this gap, we propose a novel RAG framework: ToPG (Traversal over Proposition Graphs). ToPG models its knowledge base as a heterogeneous graph of propositions, entities, and passages, effectively combining the granular fact density of propositions with graph connectivity. We leverage this structure using iterative Suggestion-Selection cycles, where the Suggestion phase enables a query-aware traversal of the graph, and the Selection phase provides LLM feedback to prune irrelevant propositions and seed the next iteration. Evaluated on three distinct QA tasks (Simple, Complex, and Abstract QA), ToPG demonstrates strong performance across both accuracy- and quality-based metrics. Overall, ToPG shows that query-aware graph traversal combined with factual granularity is a critical component for efficient structured RAG systems. ToPG is available at \url{https://github.com/idiap/ToPG}.
\end{abstract}

\section{Introduction}

\begin{figure}[ht]
  \centering
  \includegraphics[width=0.5\textwidth]{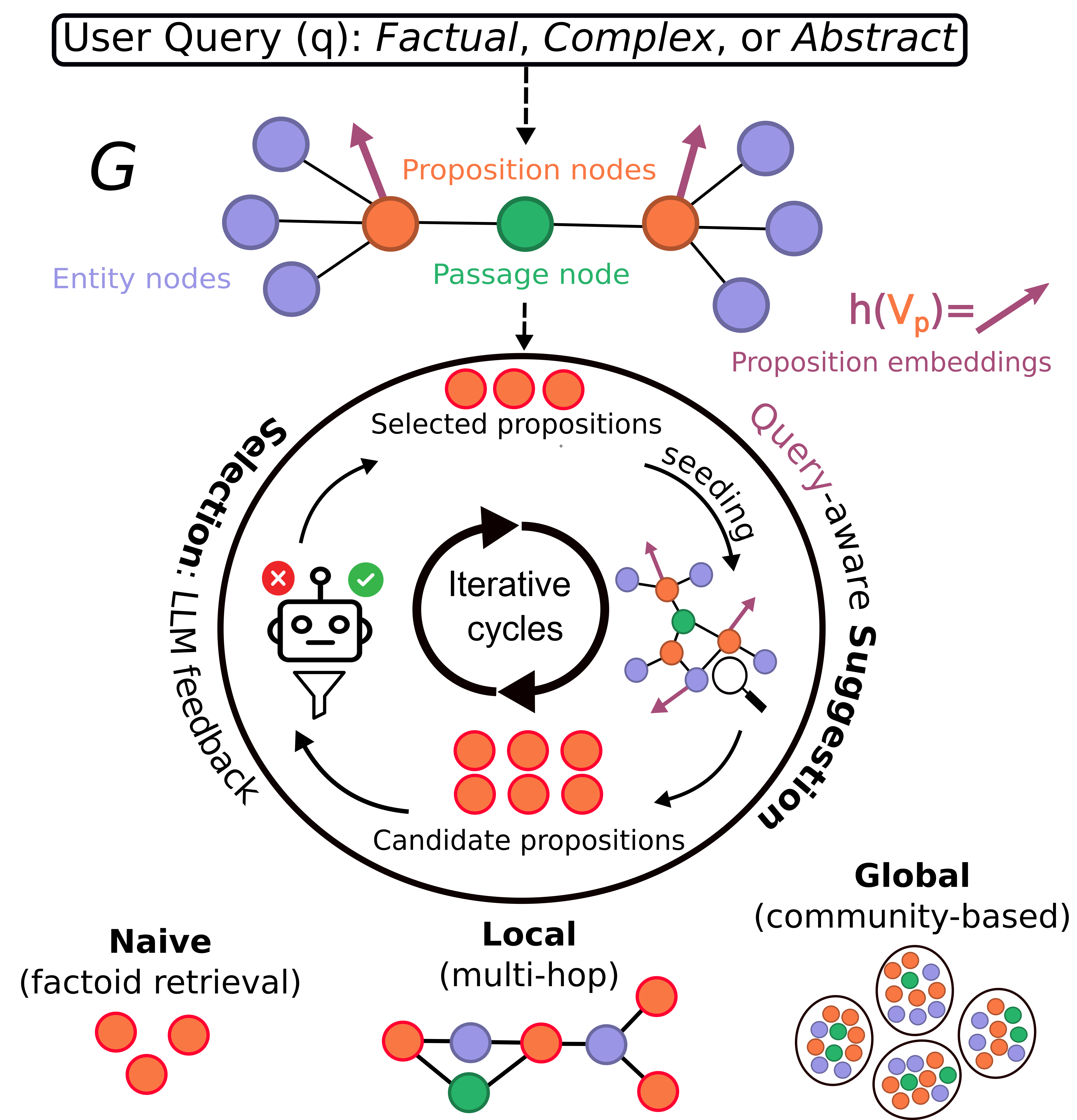}
  \caption{The ToPG framework. The system operates on a heterogeneous graph where propositions connect entities and passages. ToPG navigates this graph using iterative Suggestion-Selection cycles, allowing for three operational modes: \texttt{Naive} (factoid retrieval), \texttt{Local} (multi-hop inference), and \texttt{Global} (community-based search).}
  \label{fig:concepts}
\end{figure}

Retrieval-Augmented Generation (RAG) has become the dominant paradigm for grounding Large Language Models (LLMs). RAG directly addresses the limitations of static parametric memory, mitigating hallucinations \cite{10.1145/3703155} and improving recall, particularly for long-tail knowledge \cite{10.5555/3618408.3619049}. The standard RAG pipeline relies on Dense Passage Retrieval (DPR) over chunked documents \cite{karpukhin-etal-2020-dense}. While large embedding models (e.g., NV-Embed-v2 \cite{lee2025nvembed}) have achieved state-of-the-art performance on the MTEB benchmark \cite{muennighoff-etal-2023-mteb}, retrieval granularity represents a critical, often overlooked line for improvements. Coarse-grained passages often contain irrelevant or distracting information that degrades LLM generation \cite{10.5555/3618408.3619699}. Conversely, proposition-level retrieval (decomposing text into decontextualized atomic facts) has proven superior for direct, single-hop QA and fact checking \cite{chen-etal-2024-sub, min-etal-2023-factscore}.

Real-world complex queries often require multi-hop reasoning that necessitates connecting disparate pieces of evidence across documents. While iterative retrieval and Chain-of-Thought (CoT) approaches \cite{trivedi-etal-2023-interleaving} commonly operationalize this via successive local searches, they inherently lack a global, structured view of the corpus. To bridge this structural gap, structure-augmented RAG strategies have integrated Knowledge Graphs (KGs) \cite{zhang2025survey}. These methods explicitly model entities and relationships to support both multi-hop inference and broader, abstract queries \cite{edge2025localglobalgraphrag}.

Despite their structural advantages, current approaches face fundamental challenges. First, they lead to information loss as standard KGs enforce triples (s,p,o) representations, compressing complex text into binary relations. Second, a practical challenge exists in navigating the graph. Current strategies are broadly polarized between methods relying on purely topological heuristics (e.g., neighbours, random walks, etc.) and thus inherently ignoring edge semantics, supervised GNNs \cite{mavromatis-karypis-2025-gnn}, or LLM-driven exploration \cite{ICLR2024_10a6bdca}. To this end, we introduce ToPG (Traversal Over Proposition Graphs), a novel RAG framework that combines the granularity of propositions with \textit{query-aware graph traversal} (Figure \ref{fig:concepts}).

Unlike traditional KGs, we model the knowledge base as a heterogeneous graph of entities, propositions, and passages. This structure retains the semantic richness of atomic facts while enabling the topological connectivity of a graph. To leverage this structure, we propose a graph exploration method based on \textit{Suggestion-Selection cycles}. The Suggestion phase leverages both query similarity and graph topology to efficiently suggest new relevant propositions. The subsequent Selection phase acts as a feedback mechanism, using in-context LLM-based interpretation to prune irrelevant suggestions and seed the next iteration with high-quality evidence. To address diverse QA requirements, ToPG supports three complexity levels: \texttt{Naive} proposition retrieval, \texttt{Local} multi-hop inference, and \texttt{Global} community-based abstract QA.

\section{Methods}

\subsection{Graph Construction}

We represent the knowledge base as a heterogeneous graph $G = (V, E)$ where the node set $V = V_p \cup V_e \cup V_P$ comprises three disjoint types of nodes: atomic factual statements (propositions \(V_p\)), named entities that appear within propositions (entities \(V_e\)), and document segments that provide the source context for propositions (passages \(V_P\)). The edge set \( E = E_{p \leftrightarrow e} \cup E_{p \leftrightarrow P} \) contains two types of undirected edges, connecting each proposition to its associated entities and to the passage from which it originates.

Given a document chunk, we apply an LLM-based in-context learning function (in a few-shot setting) to sequentially extract named entities $V_e$ and propositions $V_p$. Each extracted proposition and entity node is encoded using an encoder $h(\cdot)$. Entity reconciliation is performed using cosine similarity thresholding. Details in Appendix \ref{appendix:prompt-template}.

In the resulting graph, each proposition effectively acts as a hyperedge linking multiple entities while being grounded in textual evidence\footnote{Reciprocally, entities also create hyperedges between propositions.}. Passage nodes ($V_P$) serve a structural role by connecting propositions originating from the same passage, thereby enforcing local neighborhood coherence. Unlike classical \textit{entity-centric} KGs, our representation is explicitly \textit{proposition-centric}: propositions are modeled as first-class nodes, enabling richer reasoning over factual, compositional, and multi-hop relations.

\subsection{Graph Navigation: Suggestion–Selection Cycles}

\paragraph{Suggestion-Selection Retrieval.}
We propose to navigate the graph \(G\) through Suggestion-Selection cycles. Conceptually, the suggestion step defines a function:
\begin{equation}
S_{\text{new}} = \mathrm{Suggest}_k(q, G, s_{\text{old}})
\end{equation}

Given a query \(q\), a graph $G$ and a set of already collected proposition nodes \(s_{\text{old}}\), proposes $k$ new potentially relevant nodes $S_{\text{new}}$.

An effective suggestion mechanism should account for both the semantic relevance of nodes to the query $q$, and the connectivity of nodes to the seed set $s_{old}$ in $G$. Therefore, the suggestion process should ideally be both query and graph aware. 

The Selection phase defines a function that prune irrelevant propositions from the pool $S_{\text{new}}$:
\begin{equation}
s_{\text{new}} = \mathrm{Select}(q, S_{\text{new}})
\end{equation}
It acts as feedback from the LLM and seeds the next iteration of $\mathrm{Suggest}$ by performing LLM-based relevance pruning: $\text{PROMPT}_{\text{Select}}(q, S_{\text{new}})$. By modulating the query $q$ and collected propositions $s_{old}$ during iterative Suggestion-Selection cycles, we can adapt the exploration behavior over $G$ to different question types (see section \ref{sec:naive-local-global}).

\paragraph{Query and Graph Aware Suggestions}

We introduce a retrieval strategy based on a query-aware Personalized PageRank (PPR) \cite{10.1145/511446.511513}. An intuitive example is available in Appendix \ref{appendix:QAT}. We propose to determine new candidate propositions as
\begin{equation}
S_{\text{new}} = \mathrm{top}_k\bigl(\mathrm{PPR}(M, s_{old})\bigr),
\end{equation}
where the transition matrix \(M\) combines structural and semantic information:
\begin{equation}
\begin{aligned}
M &= \mathrm{QueryAwareTransiton}(q, G, \lambda) \\
  &= \lambda T_s + (1 - \lambda) T_n.
\end{aligned}
\end{equation}
The parameter $\lambda$ controls the balance between structural and semantic guidance in $M$. The structural component $T_s$ encodes the topology of $G$: the higher the connectivity between two propositions through shared entities or passages, the greater the probability of transition between them. Thus, $T_s$ captures connectivity to the seed nodes, but, is independent of the current query $q$. In contrast, the semantic component $T_n$ maintains the same adjacency pattern as $T_s$, but weights each potential transition $(i, j)$ according to the similarity between node $j$ and the query $q$, making nodes similar to the query more attractive. Therefore, random walks are biased toward proposition nodes that are not only structurally connected to the current context $s_{old}$, but, also semantically relevant to the question. Intuitively, the resulting transition matrix $M$ encourages exploration along paths that remain consistent with the graph structure, while biased toward semantically relevant regions. Then, setting $\lambda=1$, gives a purely graph-based and query independent $\mathrm{Suggest}$ function.

More formally, \(T_s \in \mathbb{R}^{n \times n}\) is the degree-normalized transition matrix derived from the proposition--entity--passage connectivity:
\begin{equation}
T_s = \tilde{A}_{p \to eP}\, \tilde{A}_{eP \to p},
\end{equation}
where \(\tilde{A}_{p \to eP}\) and \(\tilde{A}_{eP \to p}\) denote the normalized transition matrices between propositions and entities/passages from the graph.

Then, to build $T_n$, we compute the query-based similarities \(c = \mathrm{cosine}(h(q), h(V_p))\) and apply temperature scaling and thresholding:
\begin{equation}
\tilde{c}_i =
\begin{cases}
\exp(c_i / \tau), & \text{if } c_i \ge \theta, \\
0, & \text{otherwise,}
\end{cases}
\end{equation}
where \(\tau\) is the temperature (default \(0.1\)) and \(\theta\) is the cosine threshold (default \(0.4\)). The semantic transition matrix is then defined as
\begin{equation}
T_n(i,j) =
\frac{
    \tilde{c}_j\, \mathbf{1}_{T_s(i,j) > 0}
}{
    \sum_k \tilde{c}_k\, \mathbf{1}_{T_s(i,k) > 0}
}.
\end{equation}
In both $T_s$ and $T_n$, self-connections are also canceled (eg. $T_{s}(i, i)=0$).

\paragraph{Subgraph Extraction}
As traversing the full graph \(G\) is computationally expensive, we first extract a local subgraph \(G^{*}\) using a Random Walk with Restart (hereafter named $\mathrm{GExtract}$) around the set of seeds with a target size \(l\). This process mitigates hub bias and constrains the exploration space.

\subsection{\texttt{Naive}, \texttt{Local} and \texttt{Global} modes}
\label{sec:naive-local-global}

We propose three search modes: \texttt{Naive} for simple factual queries, \texttt{Local} for complex (eg., multi-hop) queries, and \texttt{Global} for abstract questions. 

\begin{figure*}[ht]
  \centering
  \includegraphics[width=1\textwidth]{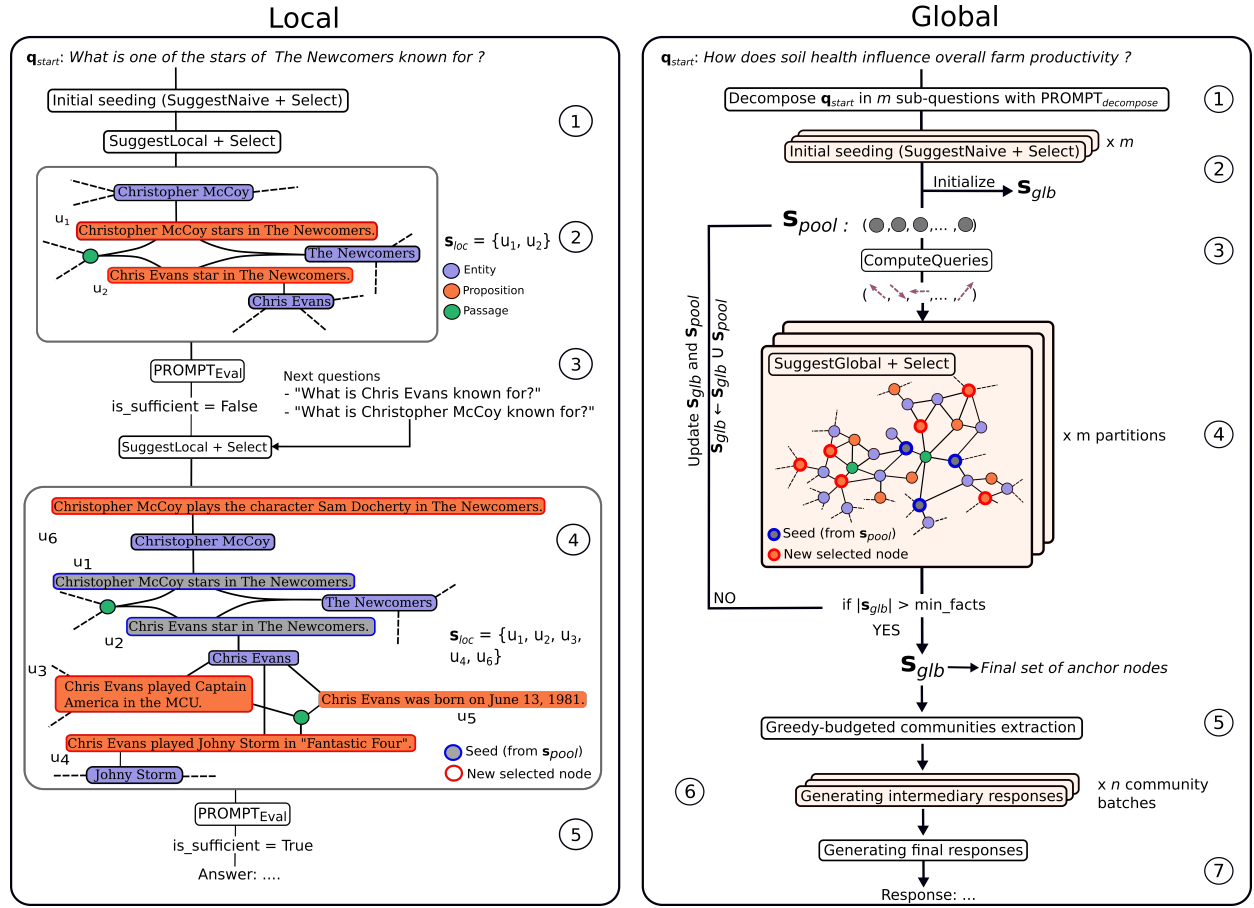}
  \caption{On the left, panel \texttt{Local} shows a step-by-step example for the \texttt{Local} mode. On the right panel, step-by-step description of the \texttt{Global} mode.}
  \label{fig:local-global}
\end{figure*}

\subsubsection{\texttt{Naive}}
Using a retrieval encoder $h$, a simple top-$k$ retrieval based on cosine similarity $S_{\text{new}} = \mathrm{top}_k\bigl(\mathrm{cosine}(h(q), h(V_p))\bigr)$ can be seen as a naive suggestion process: a retrieval over propositions that ignores both the graph $G$ and previously collected nodes $s_{\text{old}}$. We define this as $\mathrm{SuggestNaive}_k(q, G, \emptyset)$. \texttt{Naive} mode uses propositions retrieved from $\mathrm{SuggestNaive}$ as context to answer the question, bypassing the Selection step.

\subsubsection{\texttt{Local}}
\label{meth:local}
In the \texttt{Local} mode, the graph $G$ is explored through Suggestion-Selection cycles guided by LLM feedback up to a maximum number of iterations ($\texttt{max-iter}$). A step-by-step example is provided in Figure \ref{fig:local-global}\texttt{-Local}. Starting from an initial query $q_{start}$, it completes a local set of propositions \(s_{\text{loc}}=\{u_1, u_2, \text{...}, u_m \}\) that represent the evolving context for answering the query. The initial propositions are collected with $\mathrm{SuggestNaive}$ and the irrelevant ones are pruned by $\mathrm{Select}$ (step 1). This represents the initial \textit{seeding} step in $G$. 

Then, at each iteration, new candidate propositions ($S_{\text{new}}$) are proposed using $\mathrm{SuggestLocal}$, conditioned on the current query and the current context ($s_{\text{pool}}$). Candidates are then pruned by $\mathrm{Select}$. The retained propositions are added to $s_{\text{loc}}$ and will seed the next iteration. In step 2, the first iteration yields nodes $u_1$ and $u_2$.

The suggestion $S_{\text{new}} = \mathrm{SuggestLocal}_k(q, G, s_{\text{pool}})$ proceeds in three stages:
\begin{equation}
\begin{aligned}
G^{*} &= \mathrm{GExtract}(G, s_{\text{pool}}, l), \\
M &= \mathrm{QueryAwareTransiton}(q, G^{*}, \lambda), \\
\pi &= \mathrm{PPR}(M, s_{\text{pool}}), \\
S_{\text{new}} &= \mathrm{top}_k(\pi) 
\end{aligned}
\end{equation}

If the accumulated propositions remain insufficient to answer the query (determined by $\text{PROMPT}_{\text{Eval}}(q_{\text{start}}, s_{\text{loc}})$), additional targeted sub-questions are generated to guide the next iteration via $\text{PROMPT}_{\text{NextQ}}(q_{\text{start}}, s_{\text{loc}})$. In the running example (step 3), this yields two new queries, which in turn trigger a second Suggestion–Selection cycle. This cycle is seeded on $s_{\text{pool}}$, containing $u_1$ and $u_2$ (step 4). At this iteration, $S_{\text{new}}$ contained the suggested propositions: $u_3$, $u_4$, $u_5$, and $u_6$ and the $\mathrm{Select}$ call pruned $u_5$. With $s_{\text{loc}}$ now completed by $u_3$, $u_4$ and $u_6$, the query is answered in (step 5). Details and an illustrative example in Appendix \ref{appendix:local-mode}.

\subsubsection{\texttt{Global}}

While \texttt{Local} retrieval effectively handles fact-oriented or multi-hop reasoning tasks, abstract or conceptual queries \textit{"How does soil health influence overall farm productivity?"} require a broader and more diverse exploration of the graph $G$. In such cases, identifying a missing fact or reasoning chain is insufficient, where a comprehensive answer spans multiple, complementary and non-local perspectives that need to be retrieved from $G$. Rather than using $\text{PROMPT}_{\text{NextQ}}$ to predict new directions/questions as in \texttt{Local}, \texttt{Global} refines queries after each Suggestion-Selection cycle. Gathered anchor propositions $s_{glb}$, are then used to identify communities in $G$. Communities emerge naturally from the graph’s topology and can represent potential facets (i.e., individual aspects or perspectives) relevant to the query. Intermediate answers are generated from these communities, scored by relevance, and aggregated into the final response. A detailed diagram of the step-by-step process is presented in Figure \ref{fig:local-global}-Global. 

\paragraph{Steps 1-2: Seeding} The parameter $m$ controls the breadth of the exploration. We begin by decomposing the initial query $q_{\text{start}}$ into $m$ sub-queries using $\text{PROMPT}_{\text{decompose}}$. Each sub-query is sent to $\mathrm{SuggestNaive}$ and populates the pool of anchors $s_{\text{pool}}$ after irrelevant propositions are pruned with $\mathrm{Select}$. This corresponds to the \textit{seeding} step and the first anchors added to $s_{glb}$.

\paragraph{Step 3: Compute Queries} 
At each subsequent iteration, every proposition node in the current pool ($u_i \in s_{\text{pool}}$) becomes an independent exploration center, performing its own local walk through the graph. To guide these walks, we refine the query $q_i$ for each proposition using relevance feedback \cite{rocchio1971relevance}:
\begin{equation}
q_i = \alpha q_i^o + \beta q_i^+ - \gamma q_i^-.
\end{equation}
Intuitively, $q_i^o$ captures directions that previously led to $u_i$; $q_i^+$ encodes the proposition $u_i$ as a relevance signal from $\mathrm{Select}$, encouraging further exploration in this direction; and $q_i^-$ discourages directions that were previously pruned. Therefore, query vectors $q_i$ are refined by the LLM feedback provided by $\mathrm{Select}$, guiding the exploration toward promising directions while avoiding previously pruned paths (further details in Appendix \ref{appendix:compute-queries}).

\paragraph{Step 4: Iteratively explore and collect} 
To prevent the search space from growing as more facts accumulate, $s_{\text{pool}}$ is partitioned into $m$ subsets. The $\mathrm{SuggestGlobal}$ strategy operates independently on each partition $s_{\text{part}}$ with its associated queries $\mathbf{q}_{\text{part}}$. Each node $u_i$ in $s_{\text{part}}$ has its own query $q_i$ in $\mathbf{q}_{\text{part}}$. Within a partition, a subgraph $G^{*}$ is extracted around $s_{\text{part}}$, then, each proposition $u_i$ acts as a singleton seed $\{u_i\}$ with its query $q_i$, performing an individual query-aware random walk. The resulting probability distributions are aggregated within the partition, and the top-$k$ propositions are selected as new suggestions.

In summary, $\mathrm{SuggestGlobal}_k(\mathbf{q}_{\text{part}}, G, s_{\text{part}})$ is defined as:
\begin{equation}
\begin{aligned}
G^{*} &= \mathrm{GExtract}(G, s_{\text{part}}, l), \\
M_i &= \mathrm{QueryAwareTransiton}(q_i, G^{*}, \lambda), \\
\pi_i &= \mathrm{PPR}(M_i, \{u_i\}), \\
S_{\text{new}} &= \mathrm{top}_k\bigl(\sum_{i} \pi_i)
\end{aligned}
\end{equation}
The process is iterated until $\texttt{min\_facts}$ are collected (or $\texttt{max-iter}$ iterations are completed). See details in Appendix \ref{appendix:global-mode}.

\paragraph{Step 5: Identifying communities}  
The final set of collected anchor propositions $s_{\text{glb}}$ is used to extract associated communities from $G$. Communities are identified using a hierarchical Leiden algorithm \cite{traag2019louvain}. We then follow a greedy budgeted strategy: each community $c$ is assigned a score \(\frac{|nodes(c) \setminus S'|}{size(c)}\), representing how many yet-uncovered anchor propositions ($S'$) it includes relative to its size. Given a budget limit of $B$ total nodes, communities with the highest score are iteratively added to maximize coverage over anchor propositions. See details of the procedure in \ref{appendix:greedy-community}.  

\paragraph{Steps 6-7: Generating answers}  
Each community contains a heterogeneous mix of nodes: propositions, entities, and passages. Entities highlight central topics, propositions ground the key facts, and passages provide broader context and connect propositions together. Community content is divided into chunks of pre-specified token size and used to generate intermediary answers. Intermediary answers are ranked and combined into the final prompt, inserting the most relevant information at the beginning and the end (“lost-in-the-middle” effect \cite{liu-etal-2024-lost}), before generating the final answer.

\section{Experimental Setup}

We evaluate \texttt{Naive}, \texttt{Local}, and \texttt{Global} modes across complementary QA settings. Our evaluation spans: (i) \textbf{Simple QA}, testing the ability to retrieve isolated factual evidence; (ii) \textbf{Complex QA}, requiring the retrieval and composition of multiple evidence (eg., multi-hop queries); and (iii) \textbf{Abstract QA}, involving conceptual or multi-faceted queries that require broad, long-form synthesis beyond explicit facts.

\subsection{Datasets}

\paragraph{Simple QA}  
Following prior work \cite{gutiérrez2024hipporag}, we evaluated on a subset of 1,000 queries from PopQA \cite{mallen-etal-2023-trust} and included GraphRAG-Benchmark \cite{xiang2025usegraphsragcomprehensive} Task 1 (\textit{Fact Retrieval}), covering two distinct corpora: Medical, containing NCCN clinical guidelines, and Novel, a collection of pre-20th-century literary texts from Project Gutenberg.

\paragraph{Complex QA.}  
We used the 1,000-query subsets of the multi-hop QA datasets HotPotQA \cite{yang2018hotpotqa} and MusiQue \cite{trivedi-etal-2022-musique} from \citet{gutiérrez2025ragmemorynonparametriccontinual}. We also included two GraphRAG-Benchmark tasks: \textit{Complex Reasoning}, which requires chaining multiple evidence, and \textit{Contextual Summarization}, which requires synthesis of fragmented information. For them, we follow the Answer Accuracy metric \cite{xiang2025usegraphsragcomprehensive}.

\paragraph{Abstract QA.}  
To evaluate abstract queries, we follow the LightRAG setup \cite{guo-etal-2025-lightrag} and generate abstract questions on three corpora from the UltraDomain benchmark (college-level textbooks): Agriculture, Computer Science, and Legal \cite{qian2025memorag}. We compare responses on 4 dimensions with LLM-as-a-judge \cite{gu2025surveyllmasajudge}: Comprehensiveness, Diversity, Empowerment and finally Overall. See details and examples of queries in Appendix \ref{appendix:evaluation-examples}.

\subsection{Settings and Baselines}
For Simple and Complex QA, we evaluated three structure-augmented RAG baselines: GraphRAG \cite{edge2025localglobalgraphrag}, LightRAG \cite{guo-etal-2025-lightrag}, and HippoRAG 2 \cite{gutiérrez2025ragmemorynonparametriccontinual}. For both Simple and Complex QA, we used $k=20$ propositions for \texttt{Naive} and \texttt{Local} modes, assessing the latter with \texttt{max-iter}~$\in\{1,3\}$. Hyperparameters analysis can be found in Appendix \ref{appendix:fig-damping-lambda}. To isolate the benefits of proposition-level retrieval, we also include a vanilla passage-level RAG baseline that uses the same prompting configuration as the \texttt{Naive} mode.

For Abstract QA, we evaluate the \texttt{Global} mode with varying numbers of collected anchor propositions (200–1000). We compare two variants of query refinement: \texttt{Rocchio-style feedback} using $(\alpha{=}1,\ \beta{=}0.7,\ \gamma{=}0.15)$ \cite{rocchio1971relevance}, incorporating both selected and pruned propositions; and \texttt{Simple-feedback}, where $q_i$ ignores signals from $\mathrm{Select}$  $(\alpha{=}1,\ \beta{=}\gamma{=}0)$. For fairness, we evaluate against GraphRAG and LightRAG, as both explicitly support abstract-level QA with dedicated global/hybrid modes.

To ensure fair comparison, all baselines use the same embedding model (\texttt{bge-large-en-v1.5}) and the same open LLM \texttt{Gemma-3-27B} \cite{gemmateam2025gemma3technicalreport} for indexing and inference using vLLM \cite{kwon2023efficient} on one H100. For experiments on GraphRAG-Benchmark, we align with the associated protocol and used GPT-4o-mini \cite{openai2024gpt4ocard} for both indexing and inference. For additional details, please see Appendix \ref{appendix:dataset-baseline-settings}. 

\section{Results}

\begin{table*}[ht]
\centering
\begin{tabular}{lccc|cccccc}
\hline
\textbf{Method} &
\textbf{MusiQue} & \textbf{HotPotQA} & \textbf{PopQA} &
\multicolumn{3}{c}{\textbf{MEDICAL$^{\dagger}$ }} &
\multicolumn{3}{c}{\textbf{NOVEL$^{\dagger}$ }} \\
 &  &  &  & FR & CR & CS & FR & CR & CS \\
\hline
Vanilla-RAG &
19.7 / 30.6 & 52.7 / 65.5 & \underline{49.2} / \underline{62.2} &
63.7 & 57.6 & 63.7 & 58.8 & 41.4 & 50.1 \\

GraphRAG (local) &
17.8 / 26.7 & 47.3 / 60.2 & 38.1 / 52.6 &
38.6 & 47.0 & 41.9 & 49.3 & 50.9 & \textbf{64.4} \\

LightRAG (local) &
16.7 / 25.6 & 48.0 / 59.9 & 39.7 / 53.4 &
62.6 & 63.3 & 61.3 & 58.6 & 49.1 & 48.9 \\

HippoRAG 2 &
24.7 / 36.2 & 55.1 / 66.9 & 38.4 / 48.6 &
66.3 & 62.0 & 63.1 & 60.1 & 53.4 & \underline{64.1} \\

ToPG-\texttt{Naive} &
19.5 / 30.3 & 49.2 / 61.0 & \textbf{51.6 / 63.9} &
\textbf{72.9} & 68.5 & 67.7 & 67.3 & \textbf{55.6} & 63.7 \\

ToPG-\texttt{Local} (1) &
\underline{28.0} / \underline{41.1} & \underline{55.3} / \underline{67.8} & 48.4 / 59.5 &
\underline{72.5} & \underline{68.5} & \textbf{68.8} &
\underline{67.0} & \underline{55.0} & 61.2 \\

ToPG-\texttt{Local} (3) &
\textbf{34.0 / 47.0} & \textbf{59.3 / 72.7} & 48.9 / 60.2 &
72.6 & \textbf{69.2} & \underline{68.3} &
\textbf{67.6} & 53.9 & 61.0 \\

\rowcolor{gray!15}
$\Delta$ \texttt{Local} (3) - \texttt{Naive} &
$\uparrow$ 14.5 / 16.7 &
$\uparrow$ 10.1 / 11.7 &
$\downarrow$ 2.7 / 3.7 &
$\downarrow 0.3$ &
$\uparrow 0.5 $ &
$\uparrow 0.6 $ &
$\uparrow 0.3 $ &
$\downarrow 1.7 $ &
$\downarrow 2.7 $ \\
\hline
\end{tabular}
\caption{Results on Simple and Complex QA tasks, highlighting the \textbf{best} and \underline{second-best} results. Performance (Exact Match / F1) on Simple and Multi-Hop QA (Left: MusiQue, HotPotQA, PopQA) and GraphRAG-Benchmark tasks (Right: Fact Retrieval, Complex Reasoning, Contextual Summarization) measured using the Answer Accuracy metric. ToPG-\texttt{Local} (1 and 3) report results for ($\texttt{max-iter}=1$ and $3$) respectively. $\Delta$ \texttt{Local} (3) - \texttt{Naive} shows the difference between \texttt{Local} ($\texttt{max-iter}=3$) and \texttt{Naive}.}
\label{table:simple-complex-qa}
\end{table*}

\begin{figure*}[ht]
  \centering
  \includegraphics[width=1\textwidth]{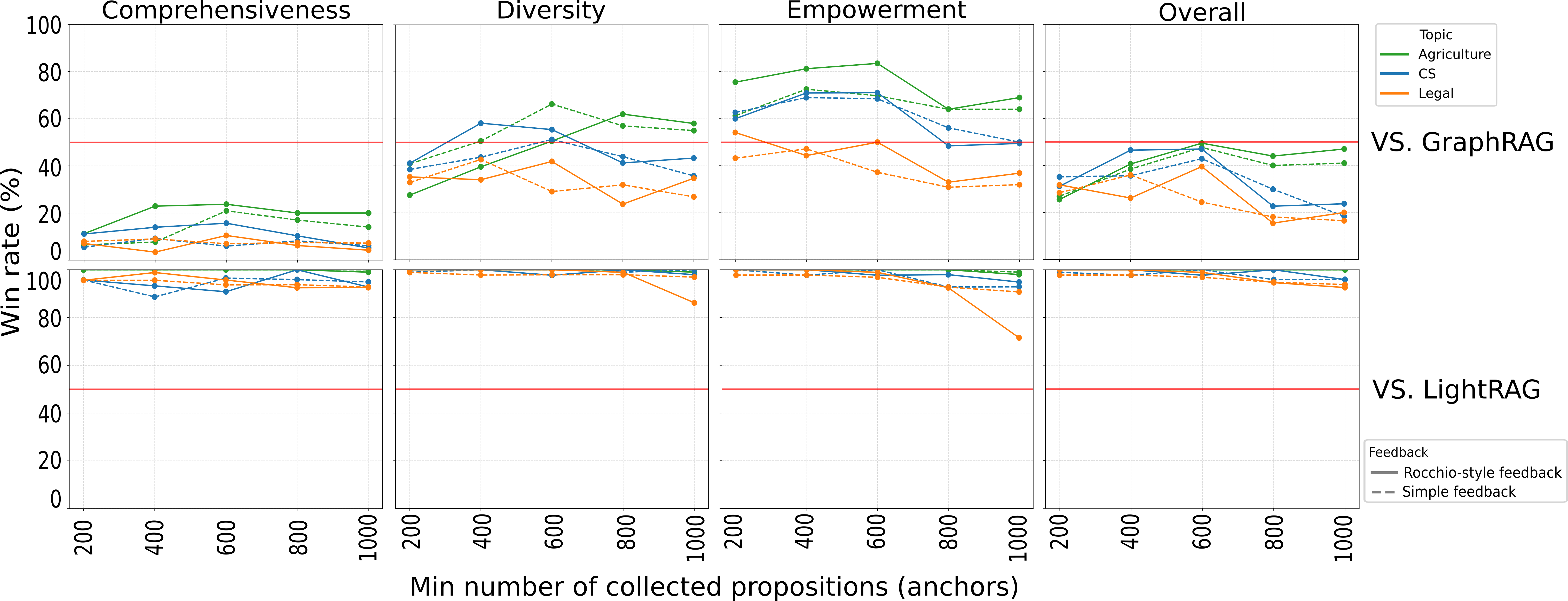}
\caption{Win rates (\%) of ToPG against GraphRAG and LightRAG across 3 corpora and 4 criteria, with increasing number of collected propositions ($200$-$1000$) and w/ or w/o Rocchio-style feedback.}
\label{fig:winrates}
\end{figure*}

\begin{figure*}[ht]
  \centering
  \includegraphics[width=1\textwidth]{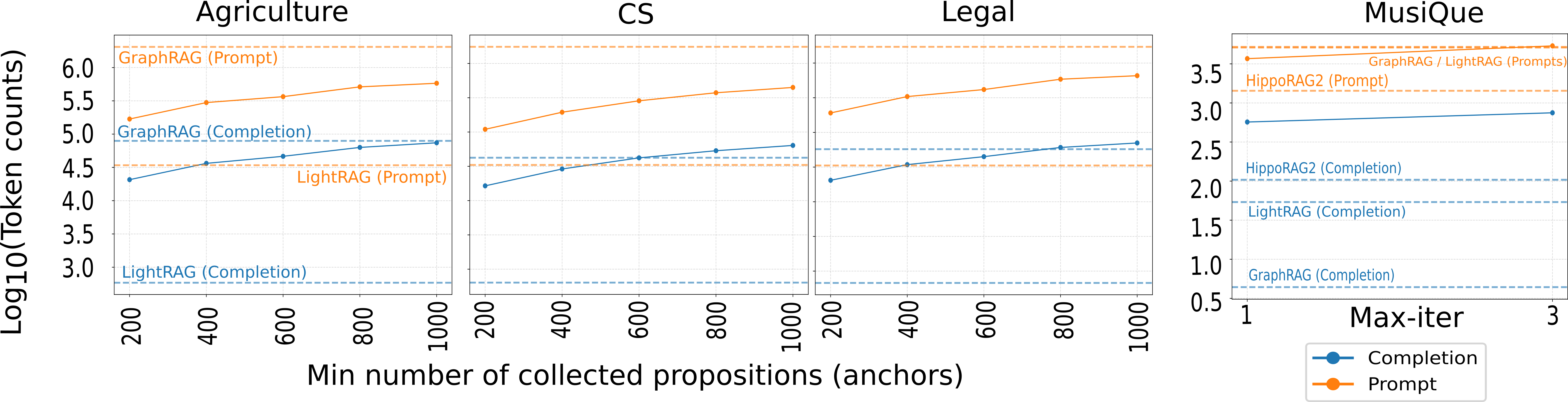}
  \caption{$\text{Log}_{10}(\text{Token-counts})$ between baselines evaluated on Agriculture, CS, Legal, MusiQue datasets.}
  \label{fig:tokens-cost}
\end{figure*}

Table \ref{table:simple-complex-qa} reports QA performance for Simple and Complex QA tasks. On Simple QA, \texttt{Naive} mode and Vanilla-RAG on passages outperform graph-based approaches by a significant margin, particularly on PopQA. However, the structural advantages of graph-based methods become apparent in Complex QA settings, notably in multi-hop scenarios where ToPG-\texttt{Local} demonstrates superior performance. Interestingly, even when configured with $\texttt{max-iter}=1$, ToPG-\texttt{Local} already exhibits significant improvements in multi-hop settings compared to its \texttt{Naive} mode. Increasing iterations to $\texttt{max-iter}=3$ yields substantial gains in multi-hop tasks but offers only marginal improvements in Complex Reasoning and Contextual Summarization (Medical and Novel corpora). For summarization tasks, both GraphRAG (local) and HippoRAG 2 also achieve competitive performance.

Figure \ref{fig:winrates} illustrates the win rates of ToPG-\texttt{Global} against baselines on Abstract QA across four criteria (see an example in Appendix \ref{appendix:abstract-qa-eval-example}). ToPG-\texttt{Global} significantly outperforms LightRAG across all configurations, reaching comparable performance with GraphRAG ($\approx 50\%$ win rate) on the Agriculture and CS datasets, though it underperforms on the Legal dataset. While GraphRAG consistently outperforms on the Comprehensiveness axis, ToPG achieves greater diversity and is perceived as more empowering in its answers. For all criteria except Comprehensiveness, increasing the number of collected facts shows a positive impact that plateaus around $600$ propositions, beyond which performance stagnates or degrades. In contrast, feedback settings for query refinement show only a negligible impact on overall performance, providing a minor improvement only in Comprehensiveness.

Figure \ref{fig:tokens-cost} compares the average token cost per abstract query, indicating that LightRAG has the lowest token cost for both input and output tokens. GraphRAG is identified with the highest token cost, particularly regarding input tokens. ToPG is cheaper than GraphRAG in completion tokens when configured with less than $600$ collected anchors, but is more costly on the MusiQue dataset.

\section{Discussion}

Graph-based approaches demonstrate competitive performance, particularly in Complex QA (multi-hop), where the graph layer effectively connects disparate named entities central to the query. Similarly to \cite{han2025ragvsgraphragsystematic}, we also note that this structural advantage, however, is often detrimental or minimal for standard factual QA, where proposition-level retrieval with ToPG-\texttt{Naive} achieves higher information density due to their self-contained and factoid content.

While baselines typically construct a standard KG with subject-predicate-object triples, their traversal often relies purely on topological heuristics (e.g., neighbours, random walks), thus neglecting the semantics encoded in the predicate. ToPG proposes a query and graph aware Suggestion mechanism to explicitly leverage the semantics of propositions, coupled with an LLM-driven Selection step that provides explicit feedback for the next iteration, but, entails the overall token cost. This Suggestion-Selection mechanism, even with only one iteration ($\texttt{max-iter}=1$), significantly improves performance over ToPG-\texttt{Naive} and alternative baselines in multi-hop settings.

In abstract QA, both GraphRAG and ToPG-\texttt{Global} rely on iterative graph exploration and exploit the inner graph modularity to extract and generate intermediary answers from node communities. While this process significantly increases token costs, it significantly improves the depth (comprehensiveness, diversity, empowerment) of generated answers over simpler keyword expansion strategies (e.g., $\text{LightRAG}$). Moreover, ToPG is designed for easier scalability and updates as it avoids pre-computing community summaries and instead uses Suggestion-Selection cycles for community exploration. Our observations suggest that the utility of collecting additional anchors is saturated by the current LLM's reasoning capacity, implying that further benefits would only arise when using a stronger base model. 

Overall, our results suggest that query-aware exploration over a graph of granular information units is the critical component, rather than the formal structure of the KG with strict predicates. Traversal of the proposed heterogeneous graph through an effective Suggestion-Selection mechanism shows robust performance and versatility across different QA tasks.

\section{Related Work}
Early strategies for complex question answering combine retrieval with reasoning via interleaving DPR with CoT or question decomposition techniques \cite{trivedi-etal-2023-interleaving, press-etal-2023-measuring, patel-etal-2022-question, shao-etal-2023-enhancing}, an approach likewise employed in the \texttt{Local} mode. Furthermore, propositions have been explored as an efficient granularity level, particularly for fact-oriented QA \cite{chen-etal-2024-dense} and claim or fact checking \cite{min-etal-2023-factscore, kamoi-etal-2023-wice}. To address the need for global structural awareness, recent approaches construct KGs directly from the corpus \cite{zhang2025survey}. These systems seed retrieval using DPR over entities or triples and then navigate the resulting graph using topological heuristics such as community detection (GraphRAG) \cite{edge2025localglobalgraphrag}, ego-network (LightRAG) \cite{guo-etal-2025-lightrag}, path search (PathRAG) \cite{chen2025pathragpruninggraphbasedretrieval}, or Personalized PageRank (HippoRAG, HippoRAG 2) \cite{gutiérrez2024hipporag, gutiérrez2025ragmemorynonparametriccontinual}. 

Combining propositions with graph structure, \citet{wang-han-2025-proprag} proposes to apply a similar approach to HippoRAG on a graph where nodes represent entities and passages, and edges link entities that co-occur within the same proposition. \citet{luo2025hypergraphrag} instead constructs a graph of propositions and perform neighborhood expansion after an initial seeding step. Unlike these approaches, ToPG leverages its Suggestion-Selection cycles and query-aware traversal to support three distinct modes tailored to different QA requirements: factoid, multi-hop, and abstract.

The Selection phase, which provides LLM-based feedback, also aligns with a broader line of work on LLM-guided KG exploration \cite{ICLR2024_10a6bdca, chen2024planongraph, ma2025thinkongraph}. These approaches typically alternate phases of search and pruning over entities and relations in the KG. Finnaly, in contrast to GraphRAG or RAPTOR \cite{sarthi2024raptor}, which rely on pre-processed summaries for abstract QA \cite{xu-etal-2022-answer, papakostas-papadopoulou-2023-model}, ToPG instead derives intermediary answers directly from the communities extracted around anchor nodes obtained through multiple Suggestion-Selection cycles.

\section{Conclusion}

ToPG reconciles fact-level granularity with graph connectivity through a heterogeneous graph composed of passages, propositions, and entities. The proposed graph navigation strategy based on iterative Suggestion-Selection cycles, while simple by design, proves highly versatile and adaptable to diverse QA requirements. The strategic modulation of the query and the collected evidence enables distinct operational modes: \texttt{Naive} (for factoid retrieval), \texttt{Local} (for complex, multi-hop reasoning), and \texttt{Global} (for abstract questions). Overall, our experiments demonstrate the efficacy of this framework and suggest that structure-augmented RAG architectures should prioritize query-aware graph traversal and factual granularity over the restrictive formal structure of traditional KGs.

\section{Limitations}

A primary limitation of our framework is the computational overhead in token cost, both during indexing and inference. Similar to other structure-augmented methods, the process of extracting propositions and building the graph significantly increases indexing costs compared to standard RAG. During inference, token consumption is inflated by the LLM-driven Selection phase (in \texttt{Local} mode) and the generation of intermediate community answers (in \texttt{Global} mode). While these mechanisms are essential for answer depth, they make ToPG less suitable for cost-critical scenarios compared to lighter alternatives like LightRAG. Future work could mitigate this by replacing the LLM selector with a specialized, lightweight classifier or by fine-tuning prompts for token efficiency.

Second, performance is also bound by the quality of the underlying graph. Relying on embedding similarity for entity disambiguation can occasionally introduce noisy or misleading edges. Furthermore, while proposition extraction enhances information density, it may result in minor information loss compared to full paragraphs. Therefore, integrating external knowledge bases (e.g., Wikipedia or DBpedia) for more robust entity linking and maintaining hybrid access to original passages could be beneficial. However, we deliberately restricted our evaluation to the proposition level for this work.

Finally, while ToPG offers three distinct operational modes (\texttt{Naive}, \texttt{Local}, \texttt{Global}), the current framework lacks an automated routing mechanism. A learned classifier capable of dynamically selecting the optimal mode based on the query complexity would make the framework more end-to-end.

\bibliography{custom}

\appendix

\section{Knowledge Base Extraction}
\label{appendix:prompt-template}

\begin{figure}[ht]
  \centering
  \includegraphics[width=0.5\textwidth]{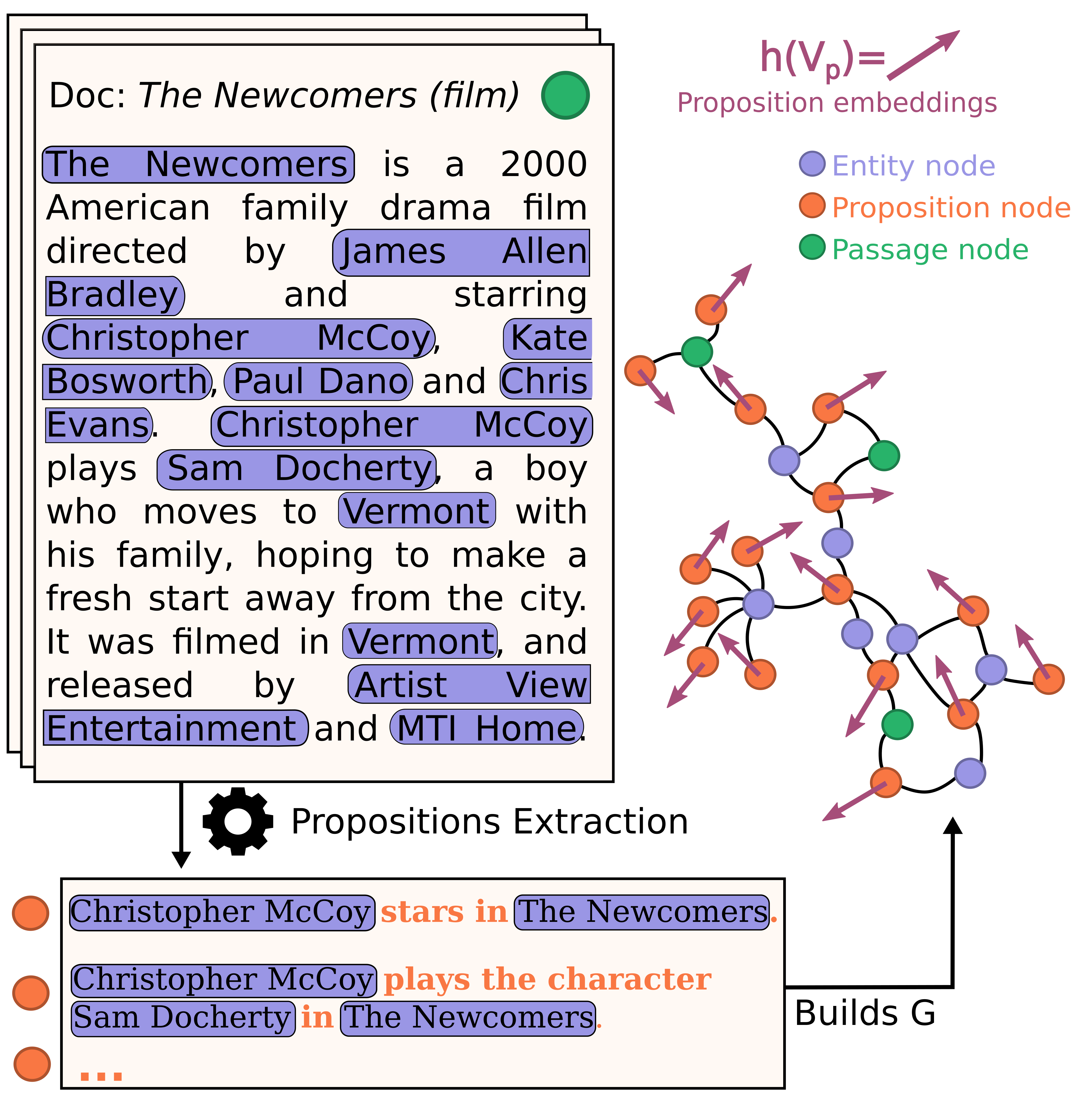}
  \caption{Knowledge base extraction process. Propositions and entities are extracted from input passages and populate the graph. Entity embeddings (used for synonym resolution) are omitted for clarity reasons.}
  \label{supp-fig:kg-construction}

\end{figure}
\begin{figure*}[!ht]
  \centering
  \includegraphics[width=0.8\textwidth]{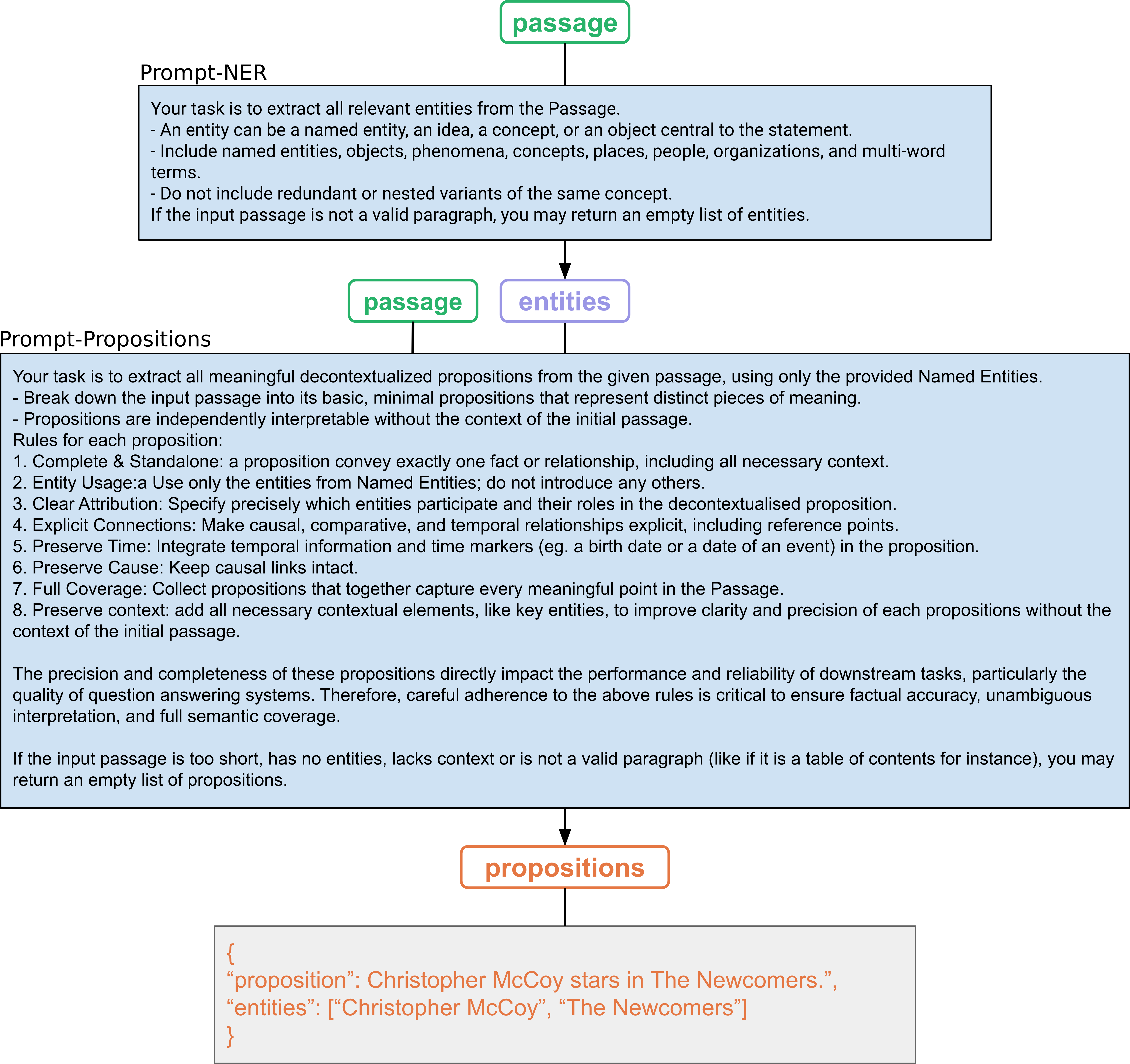}
\caption{Prompts for Named Entity Recognition and Propositions Extraction. First, named entities are extracted from the passage (\texttt{Prompt-NER}). Then, using the previously extracted entities and the original passage, propositions are extracted with \texttt{Prompt-Propositions}. Propositions are returned with their associated entities.}
\label{supp-fig:prompts-ner-propositions}
\end{figure*}

An illustration of the knowledge base extraction is presented in Figure \ref{supp-fig:kg-construction}. The prompt strategy used for entities and propositions extraction is presented in Figure \ref{supp-fig:prompts-ner-propositions}. For synonym reconciliation, given an encoder $h$, two entities $e$ and $e'$ are considered synonymous if $\mathrm{cosine}\!\big(h(e), h(e')\big) \ge \theta$ (default $0.9$).

We also report statistics for the knowledge base construction (indexing) stage of our approach. Table \ref{appendix:graph-stats-table} summarizes the resulting graph sizes, including counts of passages, propositions, and entities, as well as the total number of edges. The overall indexing cost, using the MusiQue corpus as a reference, is also provided and compared against baseline systems in Table \ref{appendix:indexing-cost-table}.

\begin{table*}[ht!]
\centering
\begin{tabular}{lrrrrrrrr}
\hline
 & MusiQue & HotPotQA & PopQA & Agriculture & CS & Legal & GB-Medical & GB-Novel \\
\hline
\# passages     & 11,704 & 9,959 & 9,101 & 9,055 & 7,337 & 16,169 & 883 & 2,400 \\
\# propositions & 83,247 & 77,409 & 73023 & 9,2840 & 58,322 & 84,134 & 8,442 & 37,868 \\
\# entities     & 82,721 & 82,909 & 79,783 & 62,341 & 31,108 & 35,732 & 3,955 & 27,071 \\
\# edges        & 350,436 & 333,799 & 309,812 & 613,688 & 320,440 & 786,049 & 49,863 & 14,9731 \\
\hline
\end{tabular}
\caption{Number of nodes (passages, propositions and entities) and edges in the graph associated with each corpora used in our experiments.}
\label{appendix:graph-stats-table}
\end{table*}

\begin{table*}[ht!]
\centering
\begin{tabular}{lrrrr}
\hline
 & ToPG & LightRAG$^\dagger$ & GraphRAG$^\dagger$ & HippoRAG 2$^\dagger$ \\
\hline
Prompts     & 70.5M & 68.5M & 115.5M & 9.2M \\
Completion  & 11.9M & 12.3M & 36.1M & 3.0M \\
\hline
\end{tabular}
\caption{Token usage comparison (prompt and completion) at indexing time for the baselines on the MuSiQue corpus (11,656 passages for 1.3M tokens).  $^\dagger$Values reported from \citet{gutiérrez2025ragmemorynonparametriccontinual}.}
\label{appendix:indexing-cost-table}
\end{table*}

\section{Supplementary Methods}

\subsection{Query Aware Transition: an illustrative example}
\label{appendix:QAT}

\begin{figure*}[!ht]
  \centering
  \includegraphics[width=0.8\textwidth]{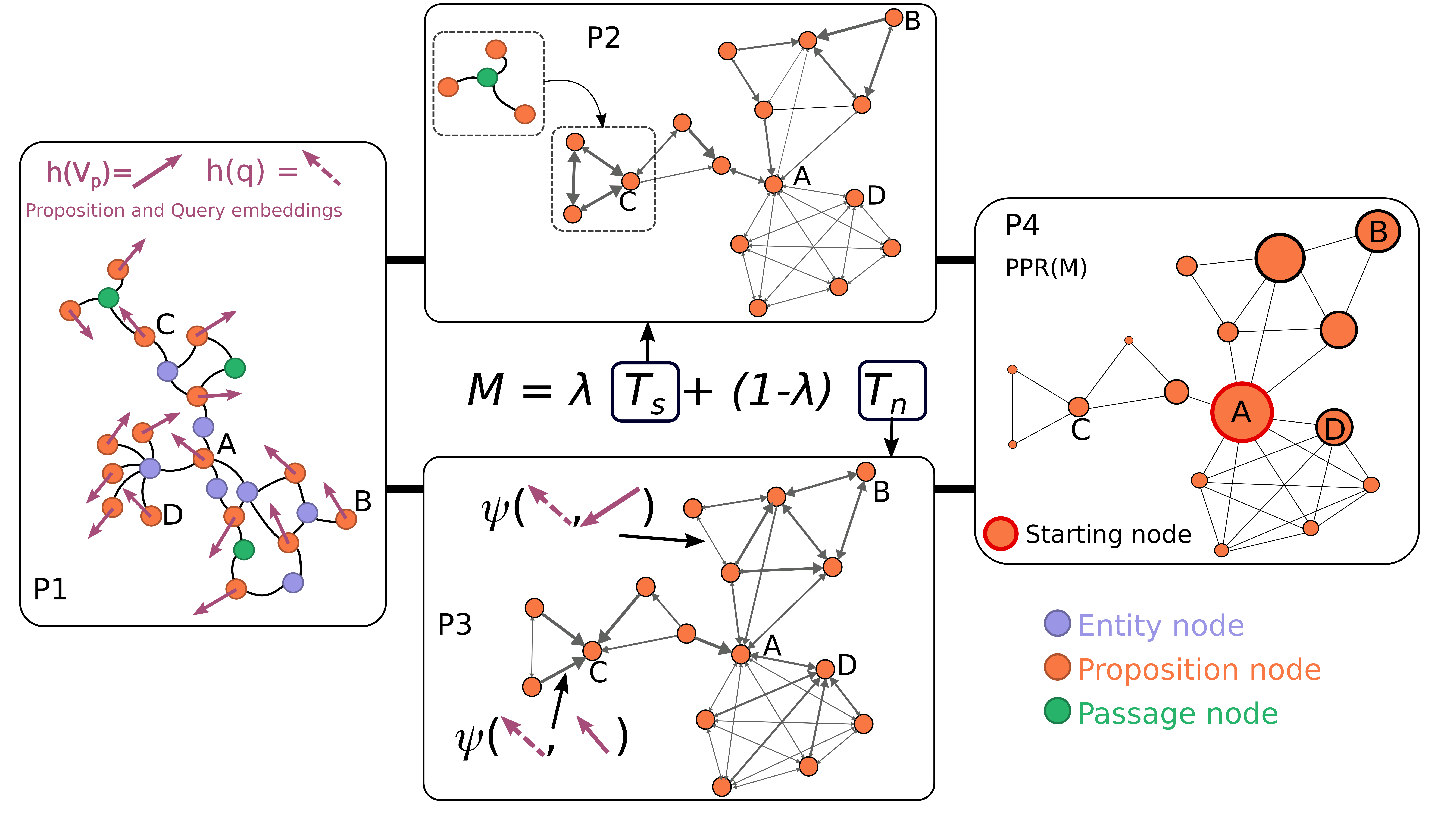}
\caption{Illustrative example of the construction and realization of the Query Aware Transitions matrix $M$ on an hypothetical subgraph $G^{*}$. $P1$ shows the subgraph $G^{*}$ with 4 landmark nodes $A$, $B$, $C$ and $D$. $P2$ and $P3$ describe the two components of $M$: $T_s$ and $T_n$. The width of the arrows is proportional to the transition probability between nodes. $P4$ illustrates the ranking obtained from the stationary distribution $\pi$ of probabilities with the PPR using $M$. The larger the node the greater the final probability and rank.}
\label{supp-fig:intuitive-method}
\end{figure*}

Figure \ref{supp-fig:intuitive-method} provides an illustrative and intuitive representation behind the construction of the query-aware transition matrix $M$. The panel $P1$ shows a hypothetical input subgraph $G^{*}$, with four annotated nodes ($A$, $B$, $C$ and $D$) that serve as reference points for the next panels. $P2$ describes the proposition-projected graph associated with $T_s$, as a \textit{propositions to propositions} graph. For instance, nodes around $D$ are all connected to the same entity in $P1$, creating a clique in the resulting projected graph in $P2$. The width of the arrows is proportional to the transition probability between two nodes, according to their connectivity (through entities and passages) in the original graph. $P3$ describes the second component of $M$: $T_n$. In this graph, the attraction of a node relative to its neighbors, indicated by the width of the arrow, is proportional to its similarity to the query (default: cosine similarity). Nodes $A$, $B$, $C$ and $D$ become attractive as their embeddings are similar to the query compared to other nodes (e.g., in the neighborhood of $D$). In $P4$, we exemplify the results of running a PPR using $A$ as the starting node and following the built $M$ transition matrix, balancing the transitions between $T_s$ and $T_n$. In this example, proposition nodes like $D$ or $B$ would be among the top-ranked nodes.

\subsection{\texttt{Local} mode}
\label{appendix:local-mode}
\begin{figure*}[h]
  \centering
  \includegraphics[width=0.8\textwidth]{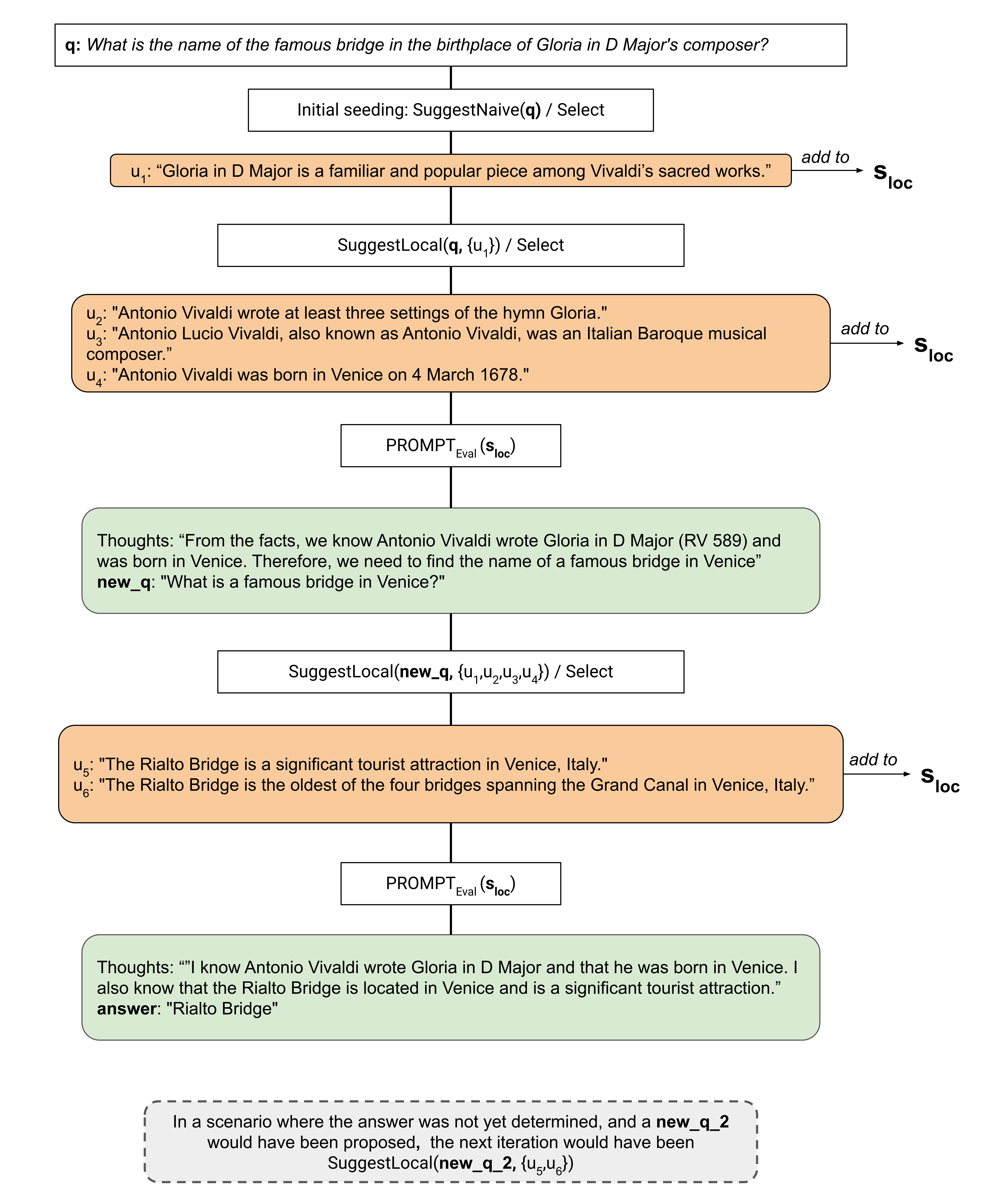}
\caption{Illustrative example of the \texttt{Local} mode.}
\label{supp-fig:example-local}
\end{figure*}

Algorithm \ref{algo:local} describes the query process in \texttt{Local} mode. An illustrative example is also provided in Figure \ref{supp-fig:example-local}.

\begin{algorithm*}[h]
\caption{Local mode with Suggestion-Selection cycles}
\label{algo:local}
\begin{algorithmic}[1]
\Require Graph \(G = (V, E)\), initial query \(q_{\text{start}}\), parameter \(\texttt{max-iter}\)
\State Initialize \(Q \gets \{q_{\text{start}}\}\)
\State \(s_{\text{loc}} \gets \emptyset\)
\State \(S_{0} \gets \mathrm{SuggestNaive}(q_{\text{start}}, \emptyset)\) \Comment{Initial seeding in the graph}
\State \(s_{\text{pool}} \gets \mathrm{Select}(q_{\text{start}}, S_{0})\)
\State \(s_{\text{pool-new}} \gets s_{\text{pool}}\)
\While{$\texttt{iteration} < \texttt{max-iter}$}
    \ForAll{\(q \in Q\)}    \Comment{Gather propositions for all questions in $s_{\text{pool-new}}$}
        \State \(S_{\text{new}} \gets \mathrm{SuggestLocal}(q, s_{\text{pool}})\) \Comment{Suggestions seeded on $s_{\text{pool}}$ and biased toward $q$}
        \State \(s_{\text{new}} \gets \mathrm{Select}(q, S_{\text{new}})\)
        \State \(s_{\text{pool-new}} \gets s_{\text{pool-new}} \cup s_{\text{new}}\)
    \EndFor
    \State \(s_{\text{loc}} \gets s_{\text{loc}} \cup s_{\text{pool-new}}\) \Comment{Complete $s_{\text{loc}}$}
    \State \(s_{\text{pool}} \gets s_{\text{pool-new}}\)
    \State \(s_{\text{pool-new}} \gets \emptyset\)
    \If{\(\text{PROMPT}_{\text{Eval}}(q_{\text{start}}, s_{\text{loc}})\) returns an answer}
        \State \Return answer
    \Else
        \State \(Q \gets \text{PROMPT}_{\text{NextQ}}(q_{\text{start}}, s_{\text{loc}})\) \Comment{Evaluate with $s_{\text{loc}}$}
    \EndIf
    \State $\texttt{iteration}++$
\EndWhile
\State \Return failure to determine answer
\end{algorithmic}
\end{algorithm*}

\subsection{\texttt{Global} mode: Compute queries}
\label{appendix:compute-queries}

The approach for query vector refinement is inspired by the Rocchio algorithm \cite{rocchio1971relevance}, which applies relevance feedback to refine a query by weighting vectors of relevant and non-relevant documents.

We adapt this principle to compute a refined query vector $q_i$ for each newly collected proposition node $u_i \in s_{\text{pool}}$. The refinement process combines: the directions that lead to $u_i$ in the previous walks, the semantic representation of $u_i$ itself, and, the directions that were pruned.

Let $\mathcal{C}_i$ denote the set of partitions where $u_i$ was identified. For a given partition $c \in \mathcal{C}_i$, let $q_{k^*}^{(c)}$ be the query vector of the walker that most likely reached $u_i$, where $k^* = \arg\max_k \pi^{(c)}_{k}(u_i)$. Furthermore, let $\bar{S}^{(c)}_{\text{new}} = S^{(c)}_{\text{new}} \setminus s^{(c)}_{\text{new}}$ be the set of candidate nodes that were pruned by the $\mathrm{Select}$ procedure in that partition and $\bar{h}\bigl(\bar{S}^{(c)}_{\text{new}}\bigr)$ their averaged embedding.

The query vector for \(u_i\) is then given by:
\begin{equation}
\begin{aligned}
q_i &=
\alpha\, q^{o}_i
+ \beta q^{+}_i
- \gamma q^{-}_i \quad \text{where,} \\
q^{o}_i &= \frac{1}{|\mathcal{C}_i|}\sum_{c \in \mathcal{C}_i} q_{k^*}^{(c)}, \\ 
\quad
q^{+}_i &= h(u_i), \\
\quad
q^{-}_i &= \frac{1}{|\mathcal{C}_i|}\sum_{c \in \mathcal{C}_i} \bar{h}\bigl(\bar{S}^{(c)}_{\text{new}}\bigr)
\end{aligned}
\end{equation}

The coefficients $\alpha, \beta, \text{ and } \gamma$ are positive weights that modulate the influence of the initial, positive, and negative feedback components, respectively.

\subsection{\texttt{Global} mode: Collecting anchor nodes}
\label{appendix:global-mode}

Algorithm \ref{algo:global} describes the iterative exploration and collection process that builds the set of anchor propositions, before community extraction in \texttt{Global} mode.

\begin{algorithm*}[h]
\caption{Anchors selection via Iterative Suggestion-Selection (\texttt{Global} mode)}
\label{algo:global}
\begin{algorithmic}[1]
\Require Graph $G$, query $q_{\text{start}}$, breadth $m$, $\texttt{max-iter}$, minimum facts $\texttt{min\_facts}$

\State $Q \gets \text{PROMPT}_{\text{decompose}}(q_{\text{start}}, m)$ \Comment{Decompose initial query into $m$ sub-queries}
\State $s_{\text{glb}} \gets \emptyset$
\State $s_{\text{pool}} \gets \emptyset$
\State $\texttt{iteration} \gets 0$

\For{each $q \in Q$}    \Comment{Initialize $s_{\text{pool}}$ with the $m$ questions}
    \State $S_{0} \gets \mathrm{SuggestNaive}(q, \emptyset)$
    \State $s_{\text{pool}} \gets s_{\text{pool}} \cup \mathrm{Select}(q, S_{0})$
\EndFor
\State $s_{\text{glb}} \gets s_{\text{pool}}$

\While{$|s_{\text{glb}}| < \texttt{min\_facts}$ \textbf{and} $\texttt{iteration} < \texttt{max-iter}$}
    \State $s_{\text{pool-new}} \gets \emptyset$
    \State $\mathbf{q_{\text{pool}}} \gets \mathrm{ComputeQueries}(s_{\text{pool}})$ \Comment{Compute queries for the new selected nodes}
    \For{each partition $s_{\text{part}}$ in  $\mathrm{Partition}(s_{\text{pool}}, m)$} \Comment{Each partition is explored independently}
        \State $S_{\text{new}} \gets \mathrm{Suggest_{Global}}(\mathbf{q}_{\text{part}}, G, s_{\text{part}})$
        \State $s_{\text{new}} \gets \mathrm{Select}(q_{\text{start}}, S_{\text{new}})$
        \State $s_{\text{pool-new}} \gets s_{\text{pool-new}} \cup s_{\text{new}}$
    \EndFor

    \State $s_{\text{glb}} \gets s_{\text{glb}} \cup s_{\text{pool-new}}$   \Comment{Complete the global $s_{\text{glb}}$ and prepare the next seeds ($s_{\text{pool}}$)}
    \State $s_{\text{pool}} \gets s_{\text{pool-new}}$
    \State $\texttt{iteration}++$
\EndWhile

\State \Return $s_{\text{glb}}$ \Comment{Final pool of collected propositions (anchors)}
\end{algorithmic}

\end{algorithm*}

\subsection{\texttt{Global mode}: Greedy community selection with budget}
\label{appendix:greedy-community}
A complete description of the greedy procedure is presented in Algorithm \ref{algo:greedy-community}. Candidate communities $c \in C$ are pre-fitlered by size (number of nodes): $10 \le |c| \le 150$ and the budget $B$ is fixed to $8000$ in the experiments.

\begin{algorithm*}[h]
\caption{Greedy Budgeted Communities Extraction}
\label{algo:greedy-community}
\begin{algorithmic}[1]
\Require $S$: anchor nodes, $B$: budget limit, $\mathcal{C}$: candidate communities
\State $b \gets 0$, $S' \gets \emptyset$, $C' \gets \emptyset$ \Comment{budget used, nodes covered, communities selected}

\While{$S' \neq S$ and $b < B$}
    \State $c^* \gets \arg\max_{c \in \mathcal{C}} \frac{| \text{nodes}(c) \setminus S'|}{\text{size}(c)}$ \Comment{best coverage / size ratio}
    \State $C' \gets C' \cup \{c^*\}$ \Comment{Update candidates, nodes covered and budget}
    \State $S' \gets S' \cup \text{nodes}(c^*)$
    \State $b \gets b + \text{size}(c^*)$
\EndWhile

\State \Return $C'$
\end{algorithmic}
\end{algorithm*}

\section{Abstracts Questions: Protocol and Examples}
\label{appendix:evaluation-examples}
We follow the procedure described by LightRAG authors\footnote{\url{https://github.com/HKUDS/LightRAG}}. To emulate a large variety of potential queries, the LLM is first instructed to generate 5 potential users with 5 related tasks for each, given a summary of the corpus. For each task, 5 questions are generated that require a high-level understanding of the corpus. Below in Figure \ref{supp-fig:agriculture-questions} is a subset of questions generated from the Agriculture corpus containing textbooks on beekeeping.

\begin{figure*}[!ht]
  \centering
  \includegraphics[width=0.8\textwidth]{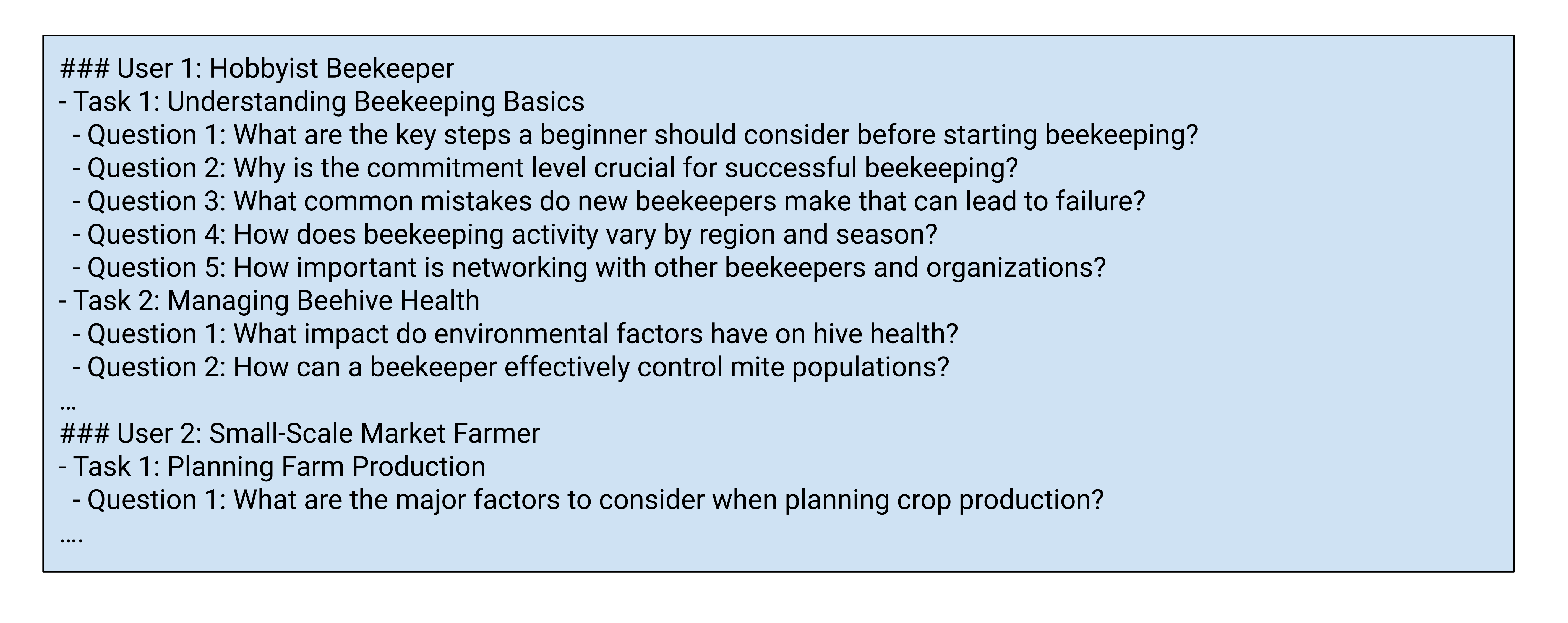}
\caption{Example of abstract questions generated for the Agriculture Corpora}
\label{supp-fig:agriculture-questions}
\end{figure*}

\section{Hyperparameters Evaluation}
\label{appendix:fig-damping-lambda}
\begin{figure}[!ht]

  \centering
  \includegraphics[width=0.5\textwidth]{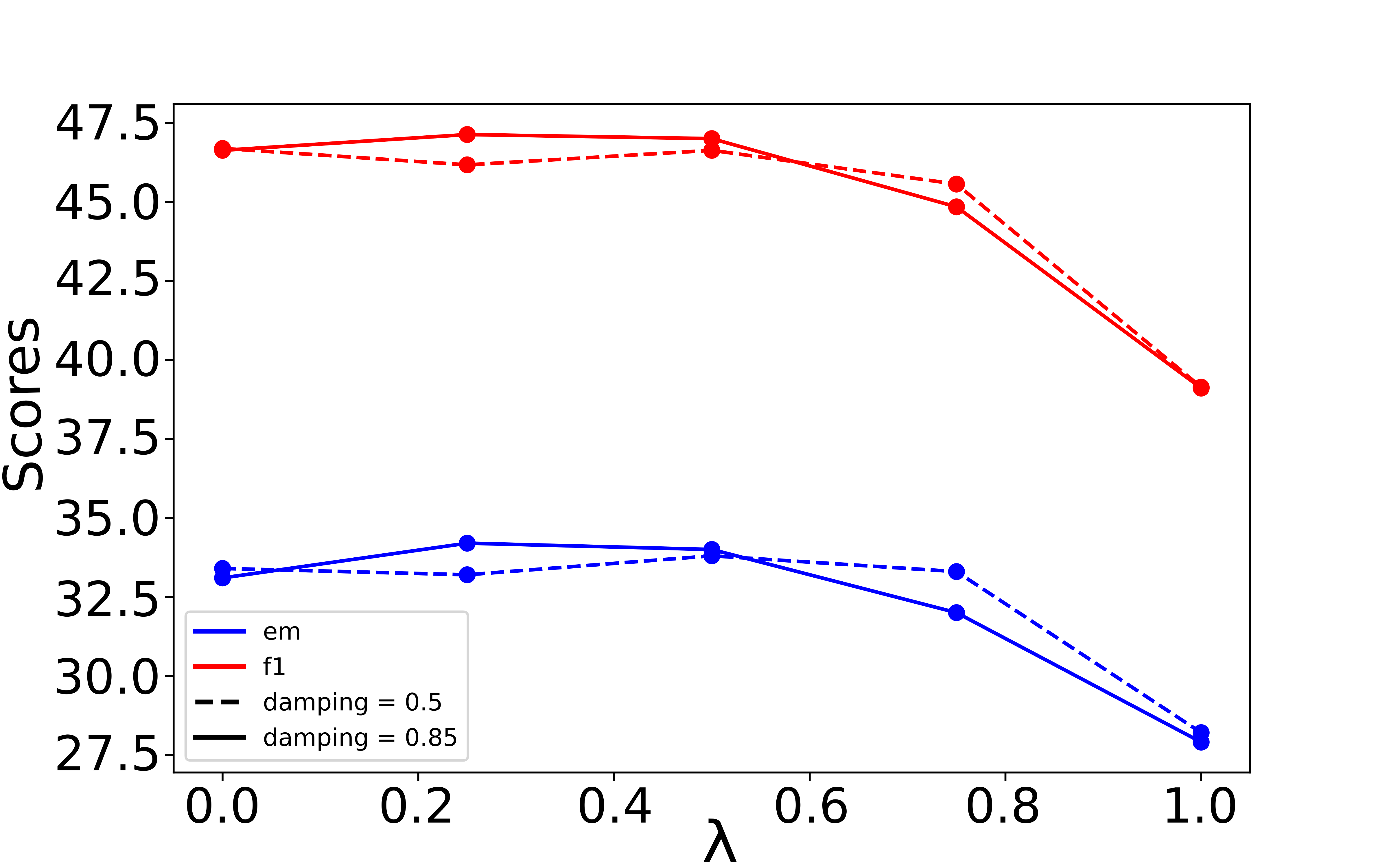}
\caption{Impact of the damping factor and $\lambda$ parameters on model performance measured by F1-score and Exact Match (EM) on MusiQue.}
\label{supp-fig:fig-damping-lambda}
\end{figure}

Figure \ref{supp-fig:fig-damping-lambda} reports the performance of different combinations of PPR damping factors and $\lambda$ values on the MusiQue dataset, using the \texttt{Local} mode with \texttt{max-iter}=3. Across both damping settings ($0.5$ and $0.85$), we observe only minor variation in EM and F1, indicating that the restart probability has limited influence on retrieval quality in these settings.

In contrast, $\lambda$, which controls the relative contribution of semantic transitions ($T_n$) versus structural transitions ($T_s$), has a pronounced impact. Performance degrades as $\lambda$ goes to $1$ and the semantic component is canceled. This highlights the importance of $T_n$ in suppressing semantically irrelevant paths when neighbors are unrelated to the query.

These observations motivated our choice of default hyperparameters: a damping factor of $0.85$ and a balanced $\lambda = 0.5$.

\section{Dataset and Baseline Details}
\label{appendix:dataset-baseline-settings}

\subsection{Dataset details}

\paragraph{Simple and Complex QA datasets}

Subsets of MusiQue (\texttt{CC-By-4.0} License), HotPotQA (\texttt{CC-By-4.0} License) and PopQA (\texttt{MIT} License) have been extracted from the repository provided by \citet{gutiérrez2025ragmemorynonparametriccontinual}\footnote{\url{https://huggingface.co/datasets/osunlp/HippoRAG_2}}. Each corpus is composed of 1,000 questions that require retrieval over one or several passages originating from Wikipedia. Additional details can also be found in Table \ref{appendix:graph-stats-table}.

\paragraph{GraphRAG-Benchmark Corpora}
We evaluate on the two corpora of GraphRAG-Benchmark\footnote{\texttt{MIT} License: \url{https://github.com/GraphRAG-Bench/GraphRAG-Benchmark}}. The Medical corpus (NCCN Guidelines) integrates data from the National Comprehensive Cancer Network (NCCN) clinical guidelines, covering diagnosis criteria, treatment protocols, and drug interactions. Additionally, the Novel corpus (Project Gutenberg) is a curated collection of pre-20th-century novels from the Project Gutenberg library. These texts exhibit complex narrative and temporal relationships. Finally, we use the same Answer Accuracy metric as used in the benchmark (see \citet{xiang2025usegraphsragcomprehensive}), which combines semantic similarity with statement-level fact checking.

\paragraph{UltraDomain Corpora}
Abstract QA uses three specialized corpora from the UltraDomain corpus\footnote{Apache 2.0 License: \url{https://huggingface.co/datasets/TommyChien/UltraDomain}}, with sizes specified in Table \ref{supp-table:ultradomain-stats}.

\begin{table*}[h]
\centering
\begin{tabular}{llc}
\toprule
\textbf{Corpus} & \textbf{Content Focus} & \textbf{Size (Tokens)} \\
\midrule
Agriculture & Beekeeping, agricultural policy, farmers, diseases and pests & 1.9M \\
Computer Science (CS) & Machine learning, data processing & 2.0M \\
Legal & Corporate finance, regulatory compliance, finance & 4.7M \\
\bottomrule
\end{tabular}
\caption{Details and size metrics for the UltraDomain corpus used in Abstract QA evaluation.}
\label{supp-table:ultradomain-stats}
\end{table*}

\paragraph{Abstract QA (LLM-as-a-Judge)}
Following \citet{guo-etal-2025-lightrag}, abstract queries are evaluated using LLM-as-a-Judge across four criteria: 
\begin{itemize}
    \item \textbf{Comprehensiveness:} How much detail does the answer provide to cover all aspects and details of the question?
    \item \textbf{Diversity:} How varied and rich is the answer in providing different perspectives and insights on the question?
    \item \textbf{Empowerment:} How well does the answer help the reader understand and make informed judgments about the topic?
    \item \textbf{Overall:} The final aggregate score combining the three criteria.
\end{itemize}

We used \texttt{Gemma-3-27B} \cite{gemmateam2025gemma3technicalreport} as the LLM during the evaluation. \texttt{Gemma-3-27B} is licensed under the \textit{Gemma Terms of Use}\footnote{\url{https://ai.google.dev/gemma/terms}}. Details on the prompts can be found on the GitHub repository at \url{https://github.com/idiap/ToPG}.

\subsection{Baseline details}

The configurations for all baseline models (HippoRAG 2, LightRAG, GraphRAG, and ToPG) are detailed in Table \ref{tab:parameters}. On the granularity level, HippoRAG 2, LightRAG, GraphRAG operate with passage-level context. LightRAG and GraphRAG additionally augment context with auxiliary KG elements (entities/relations).

For a fair comparison on the Abstract QA task, a domain-specific set of topic-related entities was defined during the indexing stage. These entities, used by GraphRAG and LightRAG, are grouped by domain:
\begin{itemize}
    \item \textbf{Agriculture:} organization, geo, event, agriculture, economic, environment.
    \item \textbf{Computer Science (CS):} organization, technology, software, metric, mathematics, hardware, computer\_science, networking.
    \item \textbf{Legal:} organization, geo, legal, regulation, financial, asset, risk, law, financial\_instrument.
\end{itemize}

We empirically found that for the Simple/Complex QA tasks, both LightRAG and GraphRAG performed optimally using their local search mode with a smaller $\texttt{top\_k}=5$ compared to standard default settings ($60$ for LightRAG and $10$ for GraphRAG). We hypothesized that the resulting large context ($\ge 8$k tokens) is detrimental for accurate factual QA with the used LLM.

To establish a strong baseline for comparison against our proposed strategy, the global mode of GraphRAG was configured with $\texttt{community\_level}=2$. While this choice significantly increased the granularity of community search, it also comes at a substantial computational cost (consuming on average $79\text{k}$ and $2.1\text{M}$ tokens for Completion and Prompts, per query.

\begin{table*}[htbp]
    \centering
    \begin{tabularx}{\textwidth}{L{0.15\textwidth} L{0.25\textwidth} L{0.4\textwidth} X}
        \toprule
        \textbf{Model} & \textbf{Task} & \textbf{Parameter} & \textbf{Value} \\
        \midrule
        \textbf{HippoRAG 2} & \textbf{Simple/Complex QA} & \texttt{top\_k} & 5 \\
        \midrule
        \multirow{6}{*}{\textbf{GraphRAG}} 
            & \multirow{6}{*}{\textbf{Simple/Complex QA}} 
                & \texttt{mode} & local \\
            & & \texttt{max\_context\_tokens} & 8000 \\
            & & \texttt{text\_unit\_prop} & 0.5 \\
            & & \texttt{community\_prop} & 0.25 \\
            & & \texttt{top\_k\_mapped\_entities} & 5 \\
            & & \texttt{top\_k\_relationships} & 5 \\
        \cmidrule{2-4}
            & \multirow{5}{*}{\textbf{Abstract QA}}
                & \texttt{mode} & local \\
            & & \texttt{community\_level} & 2 \\
            & & \texttt{use\_community\_summary} & True \\
            & & \texttt{min\_community\_rank} & 0 \\
            & & \texttt{max\_tokens} & 12000 \\
        \midrule
        \multirow{7}{*}{\textbf{LightRAG}}
            & \multirow{4}{*}{\textbf{Simple/Complex QA}}
                & \texttt{mode} & local \\
            & & \texttt{top\_k} & 5 \\
            & & \texttt{chunk\_top\_k} & 5 \\
            & & \texttt{max\_total\_tokens} & 6000 \\
        \cmidrule{2-4}
            & \multirow{4}{*}{\textbf{Abstract QA}}
                & \texttt{mode} & hybrid \\
            & & \texttt{TOP\_K} & 40 \\
            & & \texttt{CHUNK\_TOP\_K} & 10 \\
            & & \texttt{MAX\_TOTAL\_TOKENS} & 32000 \\
        \midrule
        \multirow{6}{*}{\textbf{ToPG}}
            & \multirow{6}{*}{\textbf{Simple/Complex QA}}
                & \texttt{$\lambda$} & 0.5 \\
            & & \texttt{damping $d$} & 0.85 \\
            & & \texttt{cosine\_threshold $\theta$} & 0.4 \\
            & & \texttt{subgraph $G*$ max size $l$} & 500 \\
            & & \texttt{temperature $\tau$} & 0.1 \\
            & & \texttt{top\_k} & 20 \\
        \cmidrule{2-4}
            & \multirow{4}{*}{\textbf{Abstract QA}}
                & \texttt{top\_k} & 20 \\
            & & \texttt{breadth m} & 10 \\
            & & \texttt{node budget B} & 8000 \\
            & & \texttt{min\_community\_size} & 10 \\
            & & \texttt{max\_community\_size} & 150 \\
            & & \texttt{max\_tokens\_report} & 8000 \\
            & & \texttt{passage\_token\_limit} & 500 \\
            & & \texttt{max\_tokens\_community\_chunks} & 8000 \\
        \bottomrule
    \end{tabularx}
    \caption{Parameter settings for the RAG baselines (HippoRAG 2, LightRAG, GraphRAG, and ToPG), across the evaluation tasks (Simple QA, Complex QA, and Abstract QA). Additional configuration parameters defined for ToPG include: \texttt{max\_tokens\_report} (the maximum token limit available for the final response context), \texttt{passage\_token\_limit} sets the limit for tokens used from community passages, within the \texttt{max\_tokens\_community\_chunks} available for intermediary answers. For a comprehensive description of other baseline-specific parameters, please refer to their respective documentation.}
    \label{tab:parameters}
\end{table*}

\section{Example Abstract QA and evaluation}
\label{appendix:abstract-qa-eval-example}

An example of evaluation with LLM-as-a-judge on the Agriculture corpora considering the 4 criteria (Comprehensiveness, Diversity, Empowerment and Overall) is provided in Table \ref{table:llm-as-a-judge-example}.

\begin{table*}[ht]
\centering
\begin{tabular}{p{0.22\textwidth} p{0.72\textwidth}}
\hline
\textbf{Query} &
How does soil health influence overall farm productivity? \\
\hline

\textbf{GraphRAG} &
\textbf{The Central Role of Soil Health in Farm Productivity}

Soil health is fundamentally ...

\textbf{Key Components of Healthy Soil}
Several key components contribute to overall soil health. \textbf{Organic matter} is paramount, improving soil structure....

\textbf{Agricultural Practices \& Soil Health}
Numerous agricultural practices directly impact soil health. \textbf{No-till farming} significantly improves soil health ...

\textbf{The Impact of Soil Health on Yield \& Sustainability}
Maintaining soil health is crucial for maximizing agricultural output...

\textbf{Soil Health \& Economic Viability}
Soil health is intrinsically linked to agricultural economics...

In conclusion, soil health is not merely a component ... \\
\hline

\textbf{ToPG} &
\textbf{Soil Health and Farm Productivity: A Comprehensive Report}

Improving soil health is increasingly recognized as central ...

\textbf{The Central Role of Organic Matter}
Organic matter is paramount to soil health and fertility...

\textbf{Practices for Enhancing Soil Health}
Several key practices contribute to improved soil health...

\textbf{Soil Biology and Nutrient Availability}
Healthy soil directly impacts nutrient availability for plants...

\textbf{Water Management and Soil Structure}
Soil health significantly impacts water retention and availability...

\textbf{Long-Term Sustainability and Resilience}
Improving soil health is not a complex process and can lead to cost-effective ways to feed the world and restore land...

In conclusion, prioritizing soil health is essential for maximizing farm productivity, ensuring long-term sustainability, and building resilience. \\
\hline

\textbf{LLM Decision: Comprehensiveness} &
\textbf{Winner: Answer 2}

\textbf{Explanation:} Both answers cover the core components of soil health and its impact on farm productivity. [...] Answer 2 also acknowledges challenges and considerations, like the adjustment period for transitioning to new practices, which adds to its completeness. \\
\hline

\textbf{LLM Decision: Diversity} &
\textbf{Winner: Answer 2}

\textbf{Explanation:} While both answers touch on various aspects of soil health, Answer 2 demonstrates greater diversity in its perspectives.  [...] Answer 2 also acknowledges the potential downsides of conventional practices, offering a more balanced view. \\
\hline

\textbf{LLM Decision: Empowerment} &
\textbf{Winner: Answer 2}

\textbf{Explanation:} Answer 2 is more empowering because it provides concrete examples and quantifiable data that allow the reader to understand the *magnitude* of the benefits associated with improved soil health. [...] It states that soil health is important, but doesn't provide the same level of evidence to support that claim in a way that empowers the reader to act. \\
\hline

\textbf{LLM Decision: Overall Winner} &
\textbf{Winner: Answer 2}

\textbf{Explanation:} Answer 2 is the overall winner because it excels in all three criteria. [...] While Answer 1 provides a solid overview of the topic, Answer 2 goes further in explaining the *why* and *how* of soil health, making it a more valuable and insightful resource for anyone seeking to understand this critical aspect of agricultural productivity. \\
\hline

\end{tabular}
\caption{Example of an evaluation of Abstract QA using LLM-as-a-judge, comparing answers from GraphRAG and ToPG. The model used is \texttt{Gemma-3-27B}.}
\label{table:llm-as-a-judge-example}
\end{table*}

\end{document}